# Synthesizing Reality: Leveraging the Generative AI-Powered Platform Midjourney for Construction Worker Detection

**Hongyang Zhao, Undergraduate Student,[1] Tianyu Liang, Undergraduate Student,[2] Sina Davari, Ph.D. Candidate,[3] and Daeho Kim, Ph.D.[4]**

[1]Department of Civil and Mineral Engineering, University of Toronto, 35 St. George Street, Toronto, ON, Canada; e-mail: o.zhao@mail.utoronto.ca
[2]Department of Civil and Mineral Engineering, University of Toronto, 35 St. George Street, Toronto, ON, Canada; e-mail: tianyu.liang@mail.utoronto.ca
[3]Department of Civil and Mineral Engineering, University of Toronto, 35 St. George Street, Toronto, ON, Canada; e-mail: sina.davari@mail.utoronto.ca
[4]Department of Civil and Mineral Engineering, University of Toronto, 35 St. George Street, Toronto, ON, Canada; e-mail: civdaeho.kim@utoronto.ca

## ABSTRACT

While recent advancements in deep neural networks (DNNs) have substantially enhanced visual AI's capabilities, the challenge of inadequate data diversity and volume remains, particularly in construction domain. This study presents a novel image synthesis methodology tailored for construction worker detection, leveraging the generative-AI platform Midjourney. The approach entails generating a collection of 12,000 synthetic images by formulating 3000 different prompts, with an emphasis on image realism and diversity. These images, after manual labeling, serve as a dataset for DNN training. Evaluation on a real construction image dataset yielded promising results, with the model attaining average precisions (APs) of 0.937 and 0.642 at intersection-over-union (IoU) thresholds of 0.5 and 0.5 to 0.95, respectively. Notably, the model demonstrated near-perfect performance on the synthetic dataset, achieving APs of 0.994 and 0.919 at the two mentioned thresholds. These findings reveal both the potential and weakness of generative AI in addressing DNN training data scarcity.

## INTRODUCTION

The construction industry is undergoing a gradual shift toward digitization and automation, expecting reliable and indefatigable robots to derive significant value. Robots can collect as-is construction site information (Tuhaise et al. 2023) and perform repetitive labor, allowing humans to focus on supervision and troubleshooting (Liang et al. 2021). Through digitization and robotics, the industry seeks to address chronic issues including stagnant productivity, labor shortages, and poor safety (Kim et al. 2020). To achieve these aims, construction robots are designed to operate within unstructured and dynamic environments, which demand robust object detection and situational awareness. Deep Neural Networks (DNNs), particularly in visual recognition of human workers (Baduge et al. 2022), are pivotal for this capability.

A critical issue hindering the performance of DNNs is data scarcity within the construction domain (Kim et al. 2023). In construction academia, visual datasets are generally limited to a hundred thousand images or fewer, a stark contrast to the multimillion-image datasets prevalent in other fields such as computer science (Kim et al. 2022). This data limitation becomes more acute in the case of construction sites, which, unlike typical industrial settings, feature changing contexts and significant risk to workers. Due to the expensive and time-consuming process of data collection and labeling, a data shortage prevents DNNs from achieving maximum performance potential. Additionally, a lack of diversity in training images causes visual DNN models for construction workers to be overfitted and biased, limiting scalability (An et al. 2021).

To address these issues, the authors present an approach that creates synthetic images via generative AI. For effective DNN training, whether an image is captured within physical reality is less important compared to whether it has desirable visual characteristics within a realistic context (Kim et al. 2022). By removing the need to gather images from real sites, synthetic data generation significantly reduces image-gathering costs while eliminating privacy concerns. The authors developed a dataset of synthetic images focusing on construction workers via the generative AI-powered platform, Midjourney. Onsite human workers are targeted because they are the primary players in current projects and will interact with construction robots in the future. After manual labeling, the images are used for DNN training, aiming to address the following questions: (i) is the synthetic dataset adequately diverse for scalability? and (ii) can a DNN model trained merely on synthetic images perform well in real construction settings? Answering these questions will help facilitate future use of such generative AI in DNN training.

This work demonstrates that image synthetization via generative AI can be an asset for expanding DNN-training datasets. Capable of generating a large variety of field conditions and creating a huge number of virtual images with little human effort, our approach can significantly increase the quantity and diversity of training data, which helps DNNs achieve maximum performance for worker recognition, paving the way for automation and digitization.

## BACKGROUND: EXISTING DATASETS & SYNTHETIC APPROACHES

Developing robust DNNs for visual scene understanding is contingent on the availability and quality of datasets, which provide not only learning material but also benchmarks for evaluation and improvement. This section reviews existing benchmarks in computer-vision and construction domains and analyzes previous research in synthetic data generation.

**Computer Vision & Construction Domain Benchmark Datasets.** The computer vision domain has a wide range of image benchmarks that receive regular updates to quantity and diversity. Two representative instances include ImageNet (1.3 million images; 1000 object classes including humans) (Deng et al. 2009) and Coco (330 thousand images; 91 object classes including humans) (Lin et al. 2014). While general domain datasets boast large and diverse collections, specialized datasets such as Human 3.6M (3.6 million images; human poses) (Ionescu et al. 2013) also demonstrate the size and availability of images for more specific tasks. These extensive and scalable datasets facilitate rigorous DNN model evaluations and rapid architectural advancements due to their standardization (Kim et al. 2022).

In the construction domain, datasets like the SODA "Large-Scale Small Object Detection Dataset" (20 thousand images; 15 classes including humans) (Duan et al. 2022) and the MOCS "Moving Objects in Construction Sites" benchmark by An et al. (42 thousand images; 13 classes including humans) (2021) represent valuable resources. However, when compared to the larger and more diverse datasets available in general computer vision, these are significantly smaller and lack standardization. This size and standardization gaps result in constrained DNN development, evidenced by limited network depth, increased risk of model overfitting, and slower research progress. To mitigate these issues, previous research by Kim and Chi (2019), and Bang et al. (2020), have explored post-processing methods such as image transformations and generative adversarial networks. While these methods offer benefits, they fall short in generating entirely new image contexts, thus failing to overcome the limited visual characteristics of the original datasets. The stark difference in the scale and diversity between general and construction-specific datasets highlights the need for more standardized and large construction-centric benchmarks.

**Synthetic Data Generation and Knowledge Gap.** Synthetic data is increasingly used to overcome the challenges posed by limited real-world datasets. Using Building Information Modeling (BIM) for automated labeling of real construction point clouds, Braun and Borrmann (2019) combined 4D BIM with inverse photogrammetry to produce over 30,000 automatically labeled real building elements. Assadzadeh et al. (2022), focusing on vision-based excavator pose estimation, utilized domain randomization in Unity to generate accurately annotated synthetic datasets, showcasing the method's potential in vision tasks like pose estimation. More recently, Kim et al. (2023) synthesized construction worker images via Blender, discovering that synthetic images can reduce the need for real images by 50% while enhancing DNN performance by 16%. These studies indicate the substantial role that synthetic images can play in DNN training, even in varied field conditions of construction sites.

These synthetic data generation methods share a common advantage over traditional approaches: automatic annotation, which significantly reduces human effort and cost in preparing datasets. However, these approaches also share a disadvantage: the limited realism of synthetic data. The study by Braun and Borrmann (2019) using 4D BIM with inverse photogrammetry revealed considerable deviations between synthetic results and manually labeled images with a 91.7% pixel overlap between the two data types. Excavator images rendered by Assadzadeh et al. (2022), as well as worker images produced by Kim et al. (2023), demonstrate a significant reality gap, marking a discernible discrepancy between synthetic and actual imagery. Visual inspection of these images indicates clearly noticeable differences from actual scenarios, impacting the accuracy of DNN models trained on them. While efforts to quantify this impact and enhance realism are ongoing (Ceron-Lopez et al. 2022), existing research still grapples with this challenge.

To bridge the reality gap, diffusion models have been employed. These models generate synthetic images by gradually transforming a random noise pattern into a structured image. This process involves initially generating a noisy image and then iteratively refining it to produce an image that closely matches a given text prompt or input specification (Yang et al. 2023). Diffusion models, exemplified by tools like Midjourney, can be highly accessible and synthesize images that are often indistinguishable from real ones. This technique is a novel method for creating images with completely new contexts for visual DNN training datasets, and it holds the promise of significantly narrowing the reality gap suffered by other synthetization approaches. However, despite the realistic appearance of diffusion model outputs, critical questions remain about the accuracy and performance of DNNs trained on AI-generated images versus real-world data, which this research sets out to investigate.

## METHOD: DEVELOPMENT OF SYNTHETIC TRAINING DATASET

To develop an extensive dataset of worker-related and realistic images using generative AI, the following pipeline was implemented (Figure 1). Specific parameters for prompt generation are manually established, leading to the creation of 3000 unique prompts. Subsequently, these prompts are automatically dispatched to Midjourney through Discord, producing synthetic images via diffusion. The generated images are manually labeled and subjected to training and validation using YOLOv7 (Wang et al. 2022).

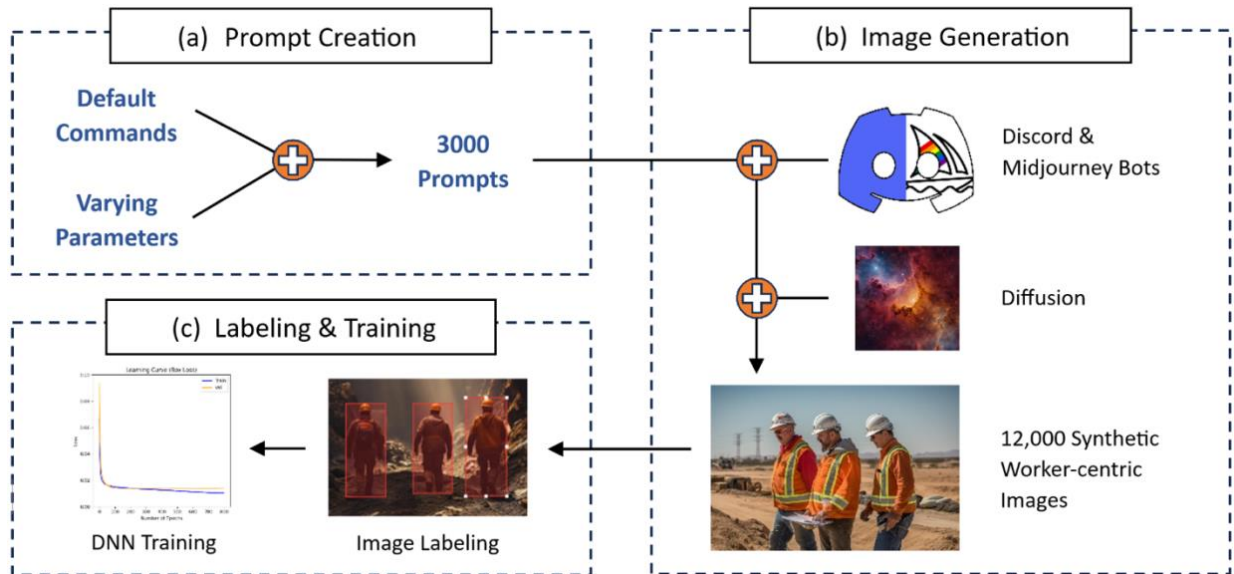

**Figure 1. The proposed synthetic dataset development pipeline**

**Prompt Creation.** A challenge of using text-to-image diffusion lies in prompt engineering. Prompts serve as the bridge between users and the AI for delivering human intentions, and the effects of their phrasing are difficult to quantify without extensive studies (Witteveen and Andrews 2022). Hence, the process of prompt creation is often considered more of an art than rigorously following steps, meaning much of our approach to prompt engineering was performed through qualitative analysis of manual trial and error. A range of parameters and the arrangement of words in the prompt were manually tested, aiming to produce realistic-looking images that varied in distance to workers, number of workers, lighting conditions, and scene contexts. The authors established a prompt format incorporating default commands and four variable parameters, as follows: /imagine prompt: three construction workers at work <location>, <weather / lighting>, <camera / film type> --ar <aspect ratio>

The phrase "/imagine prompt:" is a default command used to communicate with Midjourney. The specification of "three" most likely generates three workers, but this phrasing can produce two to five workers. The terms "construction workers" direct the generative AI to create human workers, and the terms "at work" encourage workers to appear actively engaged. Regarding the four varying parameters, <location> can drastically change background contexts and affect the sizes of the workers. <weather / lighting> impacts background contexts and plays a major role in scene illumination. <camera / film type> eliminates cartoonish results and slightly varies the color palettes of the images produced. <aspect ratio> requires the "--ar" command to specify, and it directly commands Midjourney to produce images with certain dimensions. Given the parameter variations, 3000 different combinations of prompts were created and subsequently used for image generation.

**Image Generation.** Each of the 3000 prompt combinations is sent to Midjourney (MJ) to generate four images, resulting in a total of 12,000 images. First, a Discord server is created with an MJ bot as well as a Discord bot. A Python script then sends a prompt via the Discord server to the MJ bot and waits for MJ to return an image. Typically, within 30 seconds to a minute, the MJ bot returns a single large image containing four smaller images. The Discord bot then splits this large image into the four smaller images and downloads them to a local computer. After a successful download, the Python script sends MJ the next prompt. This process is repeated until all 3000 prompts are used to produce 12,000 images (samples in Figure 2).

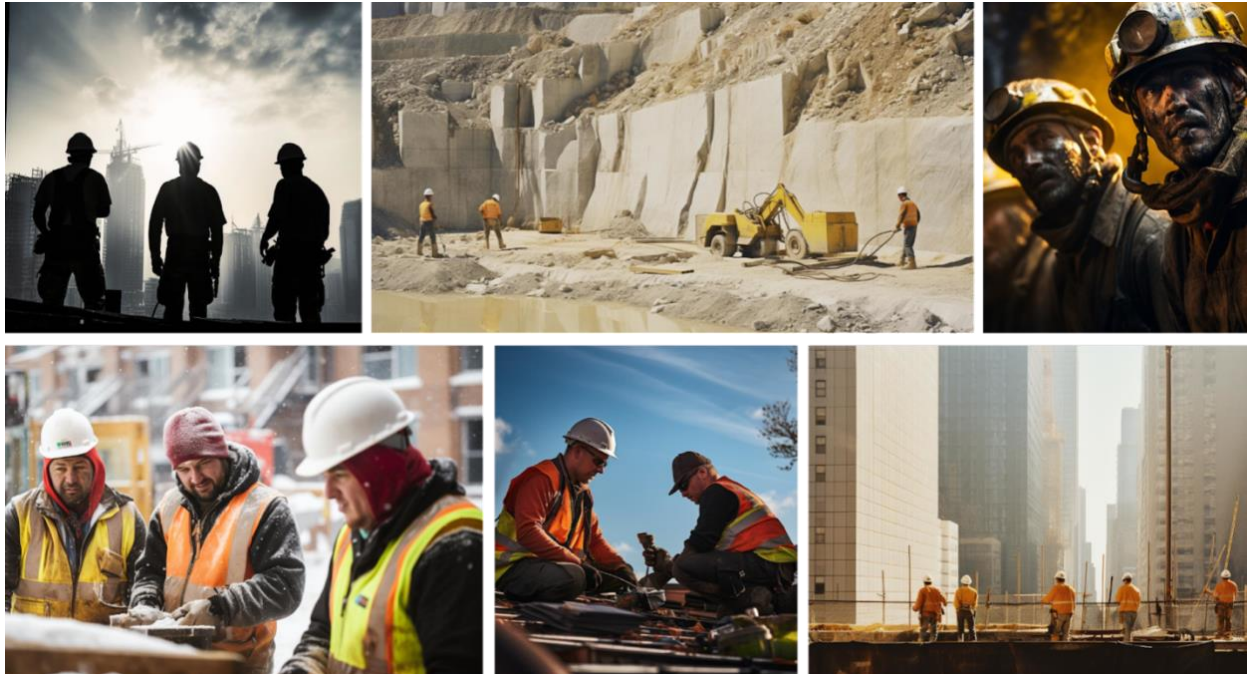

**Figure 2. Examples of the synthesized construction worker images**

**Data Labeling & DNN Model Training.** The 12,000 synthetic images are manually labeled with 2D bounding boxes with the aid of the online tool MakeSense, which draws preliminary bounding boxes around workers. This process removed eight images due to generation flaws, such as no workers in the images (samples in Figure 3), resulting in 11,992 images and 36,444 worker instances. These images are split into 9,592 for training, 1,200 for validation, and 1,200 for testing. For visual DNN model training, YOLOv7 was selected for its speed and accuracy (Wang et al. 2022). The model was trained for 800 epochs on our synthetic training set, taking 93.33 hours (about 4 days) on two NVIDIA GeForce RTX 3090 GPUs.

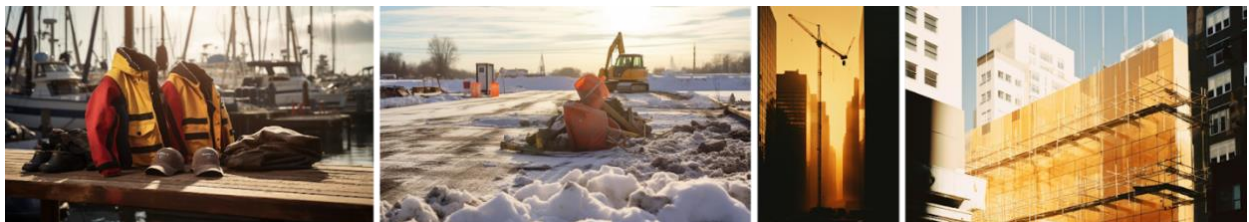

**Figure 3. Examples of erroneous synthetization results**

## RESULTS & VALIDATION

**Performance on Real Construction Dataset.** The model achieved an average precision (AP) at an Intersection over Union (IoU) threshold of 0.5 (AP0.5) of 0.937, indicating high accuracy in detecting objects when the predicted bounding boxes and the ground truth have a considerable overlap of 50% or more. Furthermore, the model reached an AP0.5-0.95 of 0.642, which is a composite measure averaging the AP calculated at different IoU thresholds ranging from 0.5 to 0.95 (in steps of 0.05). This score indicates moderate accuracy in object detection across varying degrees of overlap, reflecting the model's robustness in handling objects with different degrees of localization precision. Real-world data often contains more noise, occlusions, and variations that are challenging for the model to interpret. Nonetheless, the results indicate that the model has a degree of transferability and can perform reasonably in practical applications.

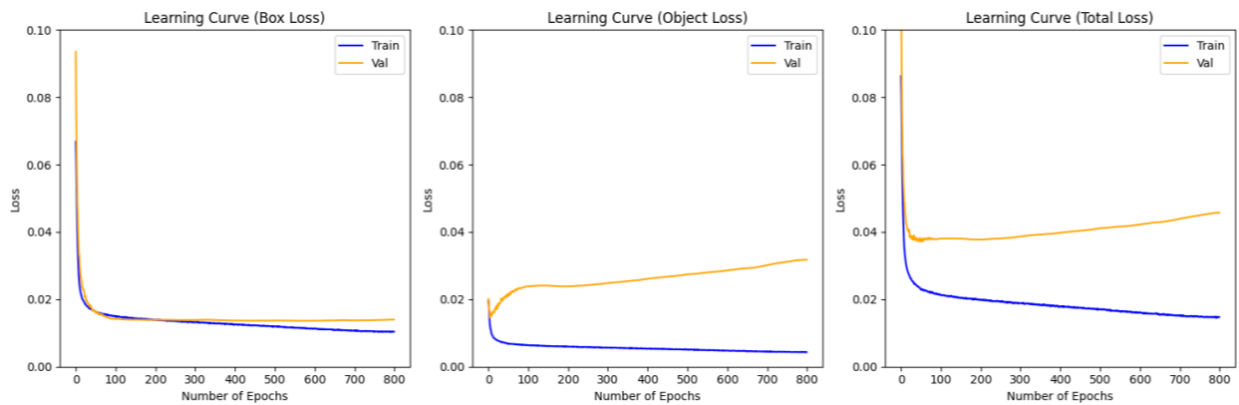

**Figure 4. Learning curve of the trained model**

In a previous study, our research team trained a YOLOv7 model with 35,017 bounding boxes on real training data from the mixture of SODA and MOCS datasets for 800 epochs and achieved an AP of 0.75 on IoU thresholds of between 0.5 to 0.95 on the same real test dataset that was utilized in this study. Upon comparing these outcomes with those derived from our model, which was trained using images synthesized by Midjourney, a notable potential in models trained on AI-generated imagery could readily be discerned.

The developed DNN model can successfully identify most of the construction workers in various scenes and provide accurate bounding boxes (left side of Figure 5). However, in certain complex scenarios, such as instances where workers are situated at a considerable distance from the camera or where significant portions of their bodies are obscured, the model's performance does encounter limitations.

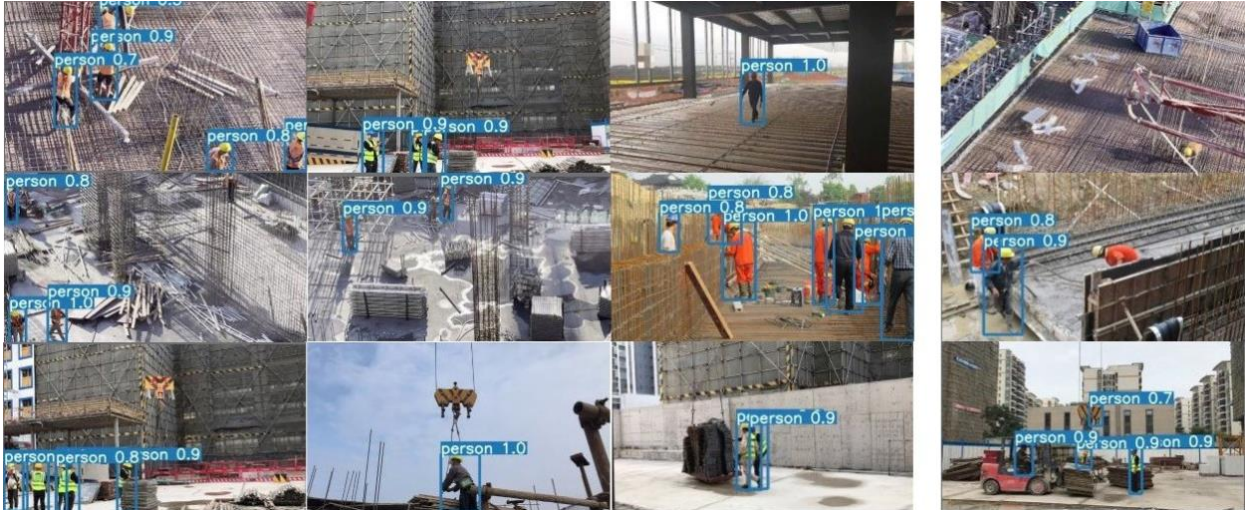

**Figure 5. Performance of our model on construction worker detection. Nine successful identifications (left) and three failure cases (right) from the real test dataset are displayed.**

**Performance on Midjourney's Synthetic Dataset.** The model achieved impressive results when evaluated on the 1200 synthetic dataset from Midjourney, with a mean average precision at an IoU threshold of 0.5 reaching 0.994, indicating near-perfect precision in detecting construction workers within these images. Furthermore, when evaluated across a range of IoU thresholds with $AP_{0.5–0.95}$, the model still performed exceptionally well, with an AP of 0.919. This level of performance suggests that the synthetic images generated by Midjourney are highly effective for training DNNs.

During validation on the synthetic dataset, the model could accurately localize construction workers in complex scenes (left side of Figure 6). However, it did occasionally encounter recognition failures in certain challenging conditions (right side of Figure 6). Overall, the model's performance affirmed the effectiveness of synthetic data for training DNNs.

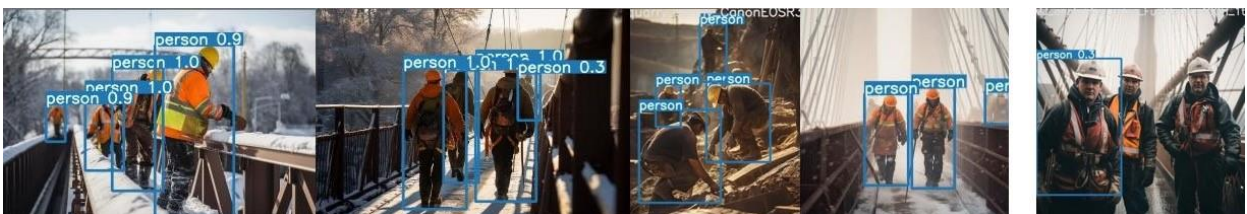

**Figure 6. Performance of our model on the test dataset created by Midjourney. Four successful identifications (left) and a failure case (right) are displayed.**

In summary, the validation results on the synthetic Midjourney dataset indicate that the developed model is well-tuned and capable of generalization. The high performance across the validation metrics suggests that synthetic data by generative AI can effectively supplement real-world data in certain stages of model training.

**DISCUSSIONS AND FUTURE WORK**

This research provides insight into using generative AI, specifically Midjourney, to create synthetic images for training DNNs in the context of worker recognition. Firstly, the experiment demonstrates the feasibility of using generative AI to create realistic and diverse images for DNN training. The synthetic images were crafted to depict a variety of construction scenarios, featuring different backgrounds, lighting conditions, camera types, and aspect ratios. This variety is crucial in training robust DNN models capable of operating in the highly variable and unstructured environments of construction sites. The results showed that the DNN model trained with synthetic images performed comparably to those trained with real-world data in certain scenarios, highlighting the potential application of generative AI in bolstering training datasets.

This study also highlights practical advantages of synthesizing images using generative AI. Our approach eliminates the need for time-consuming and costly data collection from actual construction sites, which often involves navigating privacy, confidentiality, and accessibility issues. By automating image generation, this method offers a cost-effective approach for expanding training datasets. However, while this synthesis method is innovative, it is not without limitations. Its current necessity for manual labeling, a time-consuming and labor-intensive process, introduces an element of human error. Furthermore, the synthetic nature of the images, despite their diversity and realism, may not fully encapsulate elements present in real construction sites, such as dynamic human behaviors and unique environmental changes. These limitations could potentially undermine the applicability of generative AI in supplementing DNN training datasets.

Future work will explore further integration of synthetic data into the training process, with an emphasis on refining the model to address the minute discrepancies observed in the real-world dataset performance. Addressing the logistical issue of requiring manual labeling for generative AI results would also greatly enhance the applicability of this approach. Another important direction is investigating the scalability and transferability of DNN models trained on synthetic images across different construction environments and scenarios.

**CONCLUSION**

This study investigates the potential of leveraging the generative AI-powered platform, Midjourney, for synthesizing images to train DNNs for construction worker detection. By generating 12,000 synthetic images from 3000 unique prompts, the authors aimed to alleviate data scarcity within the construction industry and enhance synthetic image realism. The findings reveal that the synthetic images significantly contribute to DNN training, offering a practical supplement to real-world datasets. However, the synthetic images still fall short of completely replicating real contexts, as the DNN model trained on real data exhibited slightly superior performance. Image synthetization via generative AI, despite creating very realistic-looking results, poses logistical drawbacks in the labeling stage compared to data synthetization within a virtual environment. Compared to automatically generated annotations in digital environments, the currently manual process of labeling generative-AI results is time-consuming and could have human error. Despite these limitations, generative AI can advance DNN training for construction scenes, contributing to more realistic and diverse datasets to enhance DNN model scalability and transferability.